\documentclass[twocolumn, 10pt]{article}

\usepackage[utf8]{inputenc}
\usepackage[a4paper, left=2cm, right=2cm, top=2cm]{geometry}

\usepackage{amsmath}
\usepackage{amssymb}

\usepackage{amsthm}
\theoremstyle{definition}
\newtheorem{definition}{Definition}

\usepackage[linesnumbered,ruled,vlined]{algorithm2e}
\usepackage{algpseudocode}

\usepackage{mathtools}

\usepackage{times}

\providecommand{\shortcite}[1]{\cite{#1}}

\author{
	Kaja Balzereit$^1$
	\and
	Oliver Niggemann$^2$
}
\date{
	$^1$ kaja.balzereit@iosb-ina.fraunhofer.de, Fraunhofer IOSB, Industrial Automation Branch (IOSB-INA), Fraunhofer Center for Machine Learning, Lemgo, Germany\\
	$^2$ oliver.niggemann@hsu-hh.de, Helmut-Schmidt University, Hamburg, Germany
}

\title{Reconfiguring Hybrid Systems Using SAT}

\begin{document}

	\maketitle

	\begin{abstract}
		Reconfiguration aims at recovering a system from a fault by automatically adapting the system configuration, such that the system goal can be reached again. 
		Classical approaches typically use a set of pre-defined faults for which corresponding recovery actions are defined manually. This is not possible for modern hybrid systems which are characterized  by frequent changes. Instead, AI-based approaches are needed which leverage on a model of the non-faulty system  and which search for a set of reconfiguration operations which will establish a valid behavior again.
		
		This work presents a novel algorithm which solves three main challenges: (i) Only a model of the non-faulty system is needed, i.e. the faulty behavior does not need to be modeled. (ii) It discretizes and reduces the search space which originally is too large---mainly due to the high number of continuous system variables and control signals. (iii) It uses a  SAT solver for propositional logic for two purposes: First, it defines the binary concept of validity. Second, it implements the search  itself---sacrificing the optimal solution for a quick identification of an arbitrary solution. It is shown that the approach is able to reconfigure faults on simulated process engineering systems.
	\end{abstract}

	\section{Introduction}
	
	Hybrid systems such as tank systems are susceptible to faults like a stuck valve or a pump failing. Usually, these faults lead to the program which controls the hybrid systems becoming invalid, i.e. the goal of the system being no longer reached \cite{blanke2006diagnosis}.
	Reconfiguration is the recovery of a system from a fault such that the system goal, usually specified by the system operator, can be reached again. Therefore, reconfiguration operations, which either exchange system components, i.e. constituents which store or process energy or material \cite{de1984qualitative}, or set inputs to a fixed value, i.e. the variables of the system that can be controlled externally, are applied. Then, an adapted program which works on the inputs that are not already set is determined to reach the system goal again. The reconfiguration operations ensure reachability of the system goal; the adapted program works on a reduced set of inputs to reach the system goal.
	
	For example in a tank system, the task of reconfiguration is not to reach an exact water level but to recover from a water level that does not allow for reaching the system goal. Therefore, some pumps and valves are set to an opened or closed position. Then, an adapted program is determined, which controls the valves and pumps, that are not set yet, to reach the system goal, e.g. an exact water level.
	
	Whilst the identification of an adapted program can be done using well-known methods from control theory or automated planning \cite{Cashmore2020} the step of reconfiguration is still not solved \cite{blanke2006diagnosis} due to some open research questions, two of which follow:
	
	\textit{RQ1: How can the reconfiguration problem be mapped onto a search problem with a reasonably sized search space?}
	Reconfiguration can be seen as search for operations setting inputs or exchanging components. Search has been extensively researched for many years now; formalizing problems as search is a common way of solving \cite{forbus1993building,pearl1987search}. Formalizing reconfiguration of hybrid systems as search enables a solution independently of the chosen modeling formalism of the hybrid system, e.g. a Hybrid Automaton or a Hybrid Bond Graph. However, search problems of hybrid systems usually have a large search space due to the high number of continuous variables having infinite domains. Hence, until now, no formalization of reconfiguration for hybrid systems exists. In this article, for the first time, we present a formalization of reconfiguration as search and show how the search space can be reduced. Therefore, continuous variables are discretized, i.e. separated into a finite number of areas, and assigned with qualitative information about validity \cite{klenk2014qualitative}.

	\textit{RQ2: How can the search for reconfiguration operations be encoded as boolean satisfiability (SAT) such that a SAT solver can be applied to solve the search problem?}
	An encoding of the search as SAT comes with two advantages: 
	(i) Reconfiguration aims at recovering from an invalid configuration, i.e. a configuration which does not allow for reaching the system goal, to a valid configuration. In contrast to quantitative approaches like differential equation systems, logic directly allows for the integration of this binary concept.
	(ii) Determining the necessary reconfiguration operations is a discrete choice leading to an explosion of combinations when the number of operations rises \cite{perlovsky1998conundrum}.
	Encodings of similar problems in boolean satisfiability have been shown to allow for intuitive modeling \cite{Feldman2020} and efficient solving \cite{metodi2014novel}. 
	To enable solving the search for reconfiguration operations using SAT, an encoding of reconfiguration in logic is needed.

	The main contribution is threefold:
	(i) For the first time, a formalization of reconfiguration for hybrid systems as search for reconfiguration operations is presented;
	(ii) An encoding of the search as boolean satisfiability in first-order logic (FOL) is shown. The use of an explicit logic allows for the representation of valid and invalid configurations and for using an efficient SAT solver;
	(iii) The approach is evaluated using representative examples from process engineering.

	The article is structured as follows: First, the related work is discussed. After that, the problem is formalized and the solution approach is presented. Then, the approach is evaluated using representative systems and a conclusion is given.

	\section{Related Work}
	
	Crow et al. \shortcite{Crow1991} have shown the analogy of reconfiguration and model-based diagnosis (MBD). MBD is often encoded in boolean satisfiability since it allows for efficiently finding minimal cardinality diagnoses \cite{metodi2014novel}. The search can be further accelerated using Multi-Operand Adders \cite{Feldman2020}.
	Such encodings allow for leveraging on many years of SAT research and have successfully been used for MBD, so our approach is based on SAT.
	
	Other approaches to reconfiguration are based on automated planning where hybrid systems are described in terms of states, actions and events \cite{Cashmore2020}. Williams et al. \shortcite{williams1996model} presented a holistic framework for the automated configuration of spacecrafts using hybrid automata and automated planning.
	However, planners are very powerful tools offering many more possibilities than required for reconfiguration, e.g. action parameters and preconditions \cite{Cashmore2020}. This comes with the drawback that creating a system and a domain model requires a large amount of expert knowledge. For reconfiguration a simpler system model is often sufficient \cite{Balzereit2020}.

	Reachability analysis of hybrid systems, which identifies if, given an initial state, a goal state can be reached, is similar to reconfiguation.
	Tools like HyTech \cite{henzinger1997hytech} and SpaceEx \cite{frehse2011spaceex} perform a symbolic analysis to determine reachability. Usual applications of such tools are safety verifications of hybrid systems: For every initial state, it is checked if the system runs out of its safety areas. However, the scope of reconfiguration is different: Reconfiguration identifies operations to restore a configuration from that a goal state is reachable. Thus, information about reachability needs to be encoded in validity of configurations.

	Qualitative Simulation (QS) is concerned with the prediction of system behavior in a qualitative way \cite{Kuipers2001}. State variables are separated into representative intervals and future intervals of the states are predicted. Even though QS is closely related to reconfiguration, it suffers from prediction uncertainties coming from the high abstraction.
	Khorasgani et al. \shortcite{khorasgani2019mode} used Minimal Structurally Overdetermined Sets for mode detection and performed residual-based fault diagnosis. 
	Our approach differs by using explicit logic, which allows for better human interpretation, and is based on SAT, enabling efficient solving \cite{Feldman2020}. 
	
	To integrate continuous data into binary logic, a discretization is done. Therefore, often intervals created by equal-width decomposition or statistics-based methods \cite{Cano2016}, leading to an improved performance of control and planning methods \cite{Konidaris2018}, are used. 
	Isermann \shortcite{Isermann2005} used filtering techniques to model the quantitative effects of a fault in a qualitative way, representing the changes to the system variables as small or large.
	However, for reconfiguration, intervals must be assigned with qualitative information allowing for the definition of validity of configurations. None of the existing methods are aimed on separating non-faulty from faulty values but instead on grouping similar data.
	
	Existing reconfiguration systems mostly handle known sets of faults with predefined reconfiguration operations \cite{gentil2004combining}. But since hybrid systems usually operate in a non-deterministic environment, predicting every possible fault is impractical due to their large number and in part rare occurrences \cite{khorasgani2019mode}. Hence, a reconfiguration method needs to determine the operations recovering validity using only observations of system variables and cannot rely on predefined instructions. 
	The novelty of our approach lies in (i) the usage of a formalization of reconfiguration as search to enable independence of the modeling formalism of the hybrid system, (ii) the reduction of the large search space by discretizing continuous variables using intervals and qualitative information, and (iii) the usage of SAT for modeling the binary validity and solving the search problem.
	
	\section{Formalization}
	
	In this section, first, reconfiguration for hybrid systems is defined. Then, a formalization of reconfiguration as search is presented (see \textit{RQ1}). Finally, a transformation of reconfiguration for hybrid automata to search is given.
	
	\subsection{Definition of Reconfiguration}
	
	Reconfiguration works on open systems, i.e. systems with inputs and outputs. Usually, the inputs are controlled by a program, e.g. a production plan or a control logic to reach the system goal \cite{iec61360}. However, the program does not handle faults such that in case of a fault the program may become invalid \cite{gentil2004combining}. Then, reconfiguration to recover the reachability of the system goal is necessary.
	
	Most of the modeling formalisms for hybrid systems have the distinction of variables into inputs and states $X$ in common \cite{de1984qualitative}:
	\begin{definition}[State] \label{def:internalState}
		A \textit{state} $x_i \in X$ evolves over time based on the input and the previous state like a water level in a tank \cite{mosterman1994behavior}.
	\end{definition} 
	In this approach, the inputs are discretized such that they can be represented by binary variables since the task is not to control a system in an optimal way but to recover validity by setting inputs to a fixed value or exchanging components. Therefore, exact numerical information about the inputs is not needed but a binary on-off control is sufficient \cite{blanke2006diagnosis}. The binary inputs are denoted in the following by $B$.
	To model observations of variables, assignments are used:
	
	\begin{definition}[Assignment]
		An \textit{assignment} $\alpha_X = \{x_1 = v_1, x_2 = v_2, ...\}, v_i \in \mathbb{R}$ maps a real value to all or some states $x_1, x_2, ... \in X$. An assignment $\alpha_B = \{b_1 = \top, b_2 = \bot, ...\}$ with $b_1, b_2, ... \in B$ is called a \textit{truth assignment}.	
	\end{definition}
	
	\begin{definition}[Configuration]
		Given an observation $\alpha_X$ to the states and an observation $\alpha_B$ to the inputs, the tuple $\langle \alpha_X, \alpha_B \rangle$ is called a \textit{configuration} of the hybrid system.
	\end{definition}
	Observability of hybrid systems, i.e. if the values of a variable are known or need to be inferred from other values, is an active research area \cite{Benazera2009}. 
	Therefore, we focus here only on reconfiguration and assume full observability.
	The task of reconfiguration is to recover a configuration from that the system goal, which is usually defined by a system operator, can be reached:
	\begin{definition}[System Goal]
		A \textit{system goal} is defined by a set $G = \{\alpha_{X_1}, \alpha_{X_2}, ... \}$ of assignments to the states $X$.
	\end{definition}
	
	\begin{definition}[Reachable]
		Given a system goal $G$ and a configuration $\langle \alpha_X, \alpha_B \rangle$, $G$ is \textit{reachable} if a program exists that converts the system to an $\alpha_{X_i} \in G$.
	\end{definition}
	
	\begin{definition}[Valid]
		Given a system goal $G$, a configuration $\langle \alpha_X, \alpha_B \rangle$ is \textit{valid} iff $G$ is reachable.
	\end{definition}
	
	\begin{definition}[Reconfiguration]
		Given an invalid configuration $\langle \alpha_X, \alpha_B \rangle$, \textit{reconfiguration} aims at finding a new assignment $\alpha_B^*$ such that $\langle \alpha_X, \alpha_B^* \rangle$ is valid., i.e. the system goal becomes reachable.
	\end{definition}
	
	\subsection{Formalization as Search}
	
	\begin{figure}
		\centering
		\includegraphics[width=\linewidth]{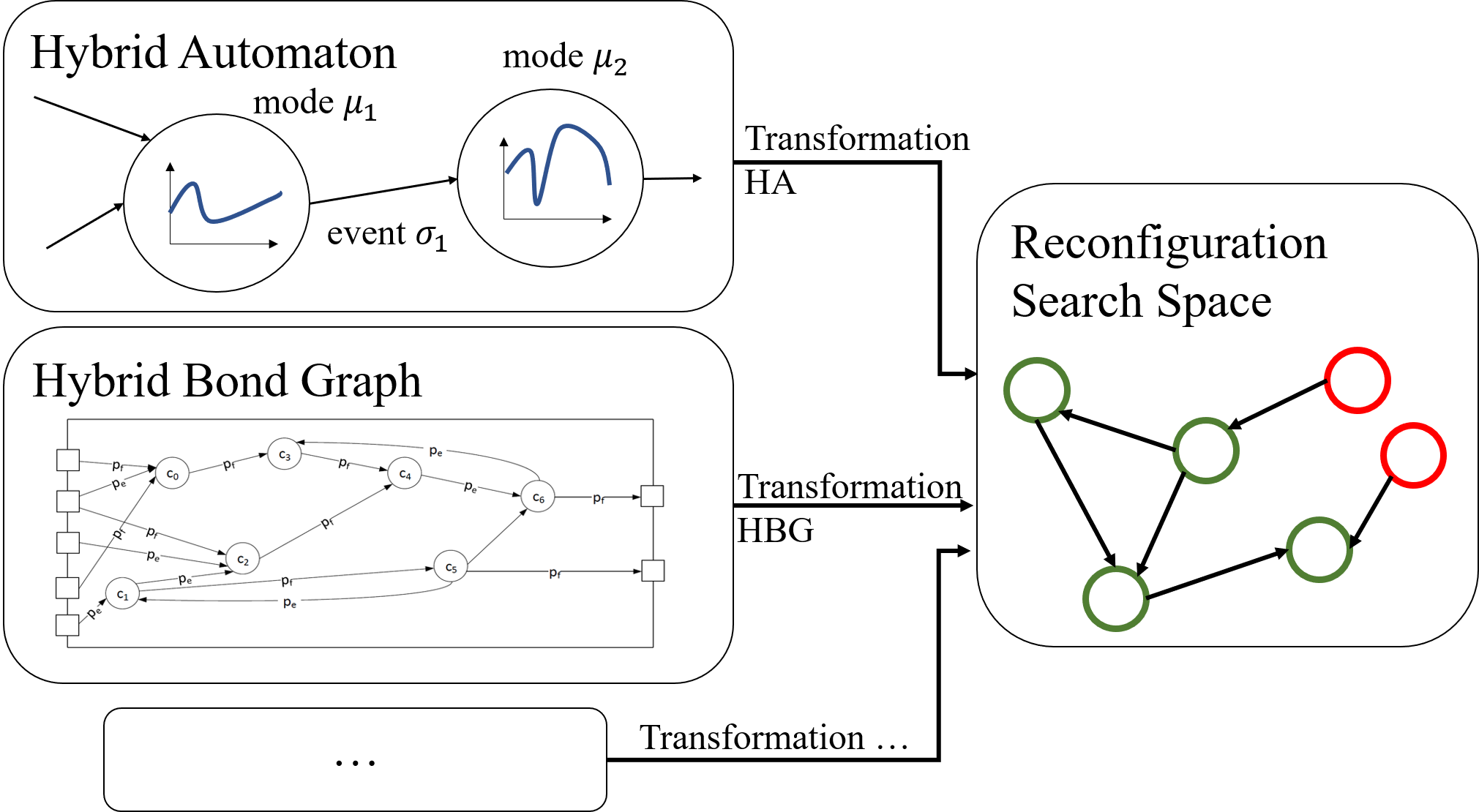}
		\caption{Based on a hybrid sytem model, e.g. a hybrid automaton or a hybrid bond graph, a specific transformation creates the reconfiguration search space. The search space contains valid (green) and invalid (red) search nodes.}
		\label{fig:reconfPlan}
	\end{figure}
	
	\begin{definition} [Search Space] \label{def:searchSpace}
		A \textit{search space} $S = \{s_1, s_2, ...\}$ consists of \textit{search nodes} $s_i$ and a set of \textit{operations} $O = \{o_k: s_i \rightarrow s_j, k \in \mathbb{N}\}$ with $s_i, s_j \in S$, which change the search node \cite{pearl1987search}.
	\end{definition}
	
	Formalizing reconfiguration as search for operations that recover a valid configuration (see Figure \ref{fig:reconfPlan}) has two advantages: (i)
	There are different ways to model hybrid systems, e.g. hybrid automata \cite{McIlraith1999,khorasgani2019mode} or hybrid bond graphs \cite{mosterman1994behavior}. The choice of modeling formalism depends on available system information. By formalizing reconfiguration as search the solution approach becomes independent of the modeling formalism. As an example, we show how the search space is created for a hybrid automaton; (ii) Search has been an active research area for many years now \cite{pearl1987search}. Sophisticated solution algorithms have been developed, which can be used by formalizing reconfiguration as search.
	For reconfiguration, the search nodes represent configurations, the operations represent the changes to the binary inputs $B$.
	
	\begin{definition} [Reconfiguration as Search] \label{def:reconfSearch}
		Given a search space $S = \{s_1, s_2, ...\}$ with $s_i = \langle \alpha_X, \alpha_B \rangle$ being configurations, an initial node $s_0$, and operations $O = \{o_k: b_i^* := \neg b_i, k \in \mathbb{N}\}$ that change the binary variables, the task of \textit{reconfiguration} is to find a set of operations $O_r \subset O$ that recovers a valid search node, so the search node after applying the operations $s^* = o_{i_r}(... o_{i_1}(s_0)), o_{i_1}, ... o_{i_r} \in O_r$ is valid.
	\end{definition}
	
	\subsection{Transformation from Hybrid Automaton to Search Space}
	
	There are many approaches to model hybrid systems:
	Cellier et al. \shortcite{cellier2013continuous} presented how the influence of inputs can be described using masks. Benazera et al. \shortcite{Benazera2009} used estimation techniques to predict the behavior of partially observable hybrid systems. 
	Blanke et al. \shortcite{blanke2006diagnosis} described hybrid systems in terms of services using expert knowledge. De Kleer et al. \shortcite{de1984qualitative} described how qualitative models of hybrid systems can be abstracted using na{\"i}ve physics.
	Figure \ref{fig:reconfPlan} shows how reconfiguration for different hybrid system models can be formalized as search. 
	As an example, the transformation for a Hybrid Automaton is shown.
	
	\begin{definition} [Hybrid Automaton (according to \cite{McIlraith1999,khorasgani2019mode})] \label{def:HA}
		A hybrid automaton is a tuple ${\cal H} = \langle {\cal M}, D, X, {\cal F}, \Sigma, \Phi \rangle$ where
		\begin{itemize}
			\item $\cal M$ is a finite set of modes $(\mu_1, \mu_2, ... )$, defined by a combination of discrete variables $D$,
			\item $X \subset \mathbb{R}^n$ defines the states,
			\item $\cal F$ is a finite set of functions $\{f_{\mu_1}, f_{\mu_2}, ...\}$ with $f_{\mu_i}: X \rightarrow X$ describing the continuous behavior in mode $\mu_i$,
			\item $\Sigma: D \rightarrow D$ is a finite set of discrete events $\{\sigma_1, \sigma_2, ... \}$ which transition the system between modes,
			\item $\Phi : \Sigma \times {\cal M} \times X \rightarrow {\cal M} \times X$ is a transition function mapping an action, a mode and a state vector into a new mode and initial state vector.
		\end{itemize}
	\end{definition}

	The inputs of the hybrid automaton are denoted by $B^{\cal H} \subset D$. Since the task of reconfiguration is not only to set inputs but also to exchange components, a set of further binary variables $B^\mathrm{EXT}$ representing if a component is exchanged is necessary. The reconfiguration operations are then represented by the possible changes to the binary variables $B = B^{\cal H} \cup B^\mathrm{EXT}$, so $o_k: b_l^* := \neg b_l~$ with $o_k \in O, b_l \in B$.
	
	The transformation consists of two steps: 
	
	(i) The search nodes $s_i = (\alpha_X, \alpha_B)$ with $B = B^{\cal H} \cup B^\mathrm{EXT}$ are formed by an observation $\alpha_{D_i}$ to the discrete variables $D \supset B^{\cal H}$ including an observation $\alpha_{B_i}$ of the binary inputs $B^{\cal H}$, and an observation $\alpha_{X_i}$ to the states $X$.
	
	(ii) To model the operations $O = \{o_k: s_i \rightarrow s_j, k \in \mathbb{N}\}$ the transition function $\Phi : \Sigma \times {\cal M} \times X \rightarrow {\cal M} \times X$ is used. Given an event $\sigma_k: B^{\cal H} \rightarrow B^{\cal H}$, and the observations $\alpha_{D_i}$, $\alpha_{X_i}$, the transition function $\Phi(\sigma_k, \alpha_{D_i}, \alpha_{X_i})$ outputs $(\alpha_{D_j}, \alpha_{X_j})$ containing a new mode with new observations to $D$ and $X$.
	The event $\sigma_k$ changing $b_l$ is mapped on an operation $o_k: b_l^* := \neg b_l$. $(\alpha_{X_i}, \alpha_{B_i})$ (from $\alpha_{D_i}$) form the initial node $s_i$ whilst $(\alpha_{X_j}, \alpha_{B_j})$ form the resulting node $s_j$.
	
	Exchanging a component $c$, i.e. setting a binary variable $b_c^\mathrm{EXT} \in B^\mathrm{EXT}$ to true, does not change the discrete variables $D$ of the hybrid automaton. It is assumed, that the states corresponding to the component are then reconfigured in a valid area.
	Thus, given an operation $o_k: b_c^{EXT*} := \neg b_c^\mathrm{EXT}$, any mode and an observation to the states, the output is the same mode and a new observation to the states. 
	
	\section{Solution Approach}
	
	There are different ways to solve search problems \cite{pearl1987search}. A commonly used algorithm is Best-First-Search (BFS) which heuristically goes through the search space for the best solution. However, BFS requires a definition of a heuristic value for each search node. Logic on the other hand allows for modeling the binary concept of valid and invalid configurations, handles the combinatorial explosion coming from the exponential rise of combinations \cite{perlovsky1998conundrum}, and allows for intuitive modeling and understandable problem representations \cite{mccarthy1981some}. This comes with the drawback, that a logic-based algorithm does not differentiate between solutions, so an arbitrary valid solution is returned whilst BFS is able to find the best solution.
	
	Our solution approach is based on encoding the search for reconfiguration operations (see Definition \ref{def:reconfSearch}) as boolean satisfiability in FOL (see \textit{RQ2}) \cite{metodi2014novel}.
	
	\begin{definition}[Reconfiguration System Model] \label{def:reconfFOL}	
		A first-order (FO) formula $SM$ represents the reconfiguration search problem (see Definition \ref{def:reconfSearch}) if
		\begin{align}
			SM &\wedge s_i \not\models \bot  \text{ if $s_i$ is valid, } \label{eq:valid} \tag{C1}\\
			SM & \wedge s_i \models \bot \text{ if $s_i$ is invalid, } \label{eq:invalid} \tag{C2}\\
			s_i &\rightarrow \varphi_i(O) \label{eq:operations} \tag{C3}
		\end{align}
		where $\varphi_i(O)$ is a formula over the operations $O$ that are available in $s_i$, holds. We call the formula $SM$ \textit{system model}. \eqref{eq:valid}, \eqref{eq:invalid} model the validity of search nodes, \eqref{eq:operations} models which operations can be applied in $s_i$ to recover validity.
	\end{definition}
	
\begin{figure}
		\centering
		\includegraphics[width=\linewidth]{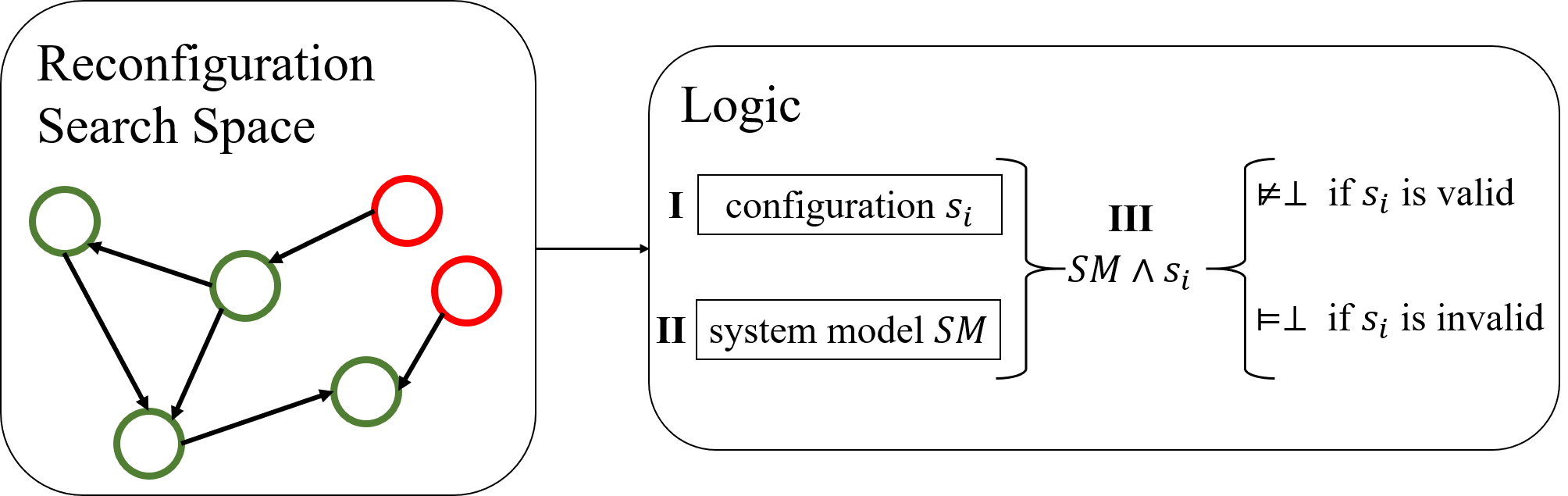}
		\caption{Steps of the solution approach.}
		\label{fig:basicIdea}
	\end{figure}

	The encoding as boolean satisfiability requires three steps (see Figure \ref{fig:basicIdea}): Since the states $X$ in general are continuous, first (step \textbf{I}) a discretization that allows for assigning qualitative information is needed. Therefore, the infinite domains are reduced to a finite number of intervals (see \textit{RQ1}). These intervals are assigned with an ordinal relation encoding if the value is okay, too high or too low (section \ref{sec:intervals}). This simplification is suitable since valid states in real-world systems often are expressed in terms of valid intervals \cite{iec61360}. In contrast to QS, where continuous variables are separated into intervals and gradient information is used to predict future states \cite{Kuipers2001}, the intervals here are not used for simulation but for the indication of valid and invalid areas.
	
	In step \textbf{II} (section \ref{sec:systemModel}) the system model $SM$ consisting of a FO formula is described. As required in Definition \ref{def:reconfFOL}, this model defines the validity of the search nodes and defines the set of operations to change the search nodes, i.e. the possible changes to the binary variables \cite{Crow1991}.
	
	In step \textbf{III} the discretized configuration (step \textbf{I}) and the system model (step \textbf{II}) are used to identify the operations to recover validity. Therefore, a SAT solver which searches for a satisfying assignment to the binary variables is used. Thus, a new input $\alpha_B^*$ is determined which recovers a valid configuration, if existing.
	
	\subsection{Discretization of Continuous Variables} \label{sec:intervals}
	
	For reconfiguration, the binary concept of validity needs to be harmonized with continuous states. Therefore, a discretization grouping the states in a suitable way to differentiate between valid and invalid configurations is necessary. Formally, a discretization is a function $g(x_1, x_2, ...)$ mapping a truth value to predicates indicating different areas.
	To identify these area, an exhaustive analysis using reachability tools like HyTech \cite{henzinger1997hytech} or SpaceEx \cite{frehse2011spaceex} might be done. However, in real-world systems, intervals on states, indicating if a variable is too low, okay or too high ($low(\cdot), ok(\cdot), high(\cdot)$), usually define areas from that a goal state is reachable \cite{iec61360,Vaennman2007}. Hence, the presented approach works on an interval decomposition of the states. However, the approach is also compatible with any discretization $g(x_1, x_2, ...)$, in case interval decomposition is not sufficient. 
	
	Many approaches to the decomposition of continuous variables into intervals have been developed. 
	However, the scope of these is not on reconfiguration for which intervals need to be assigned with qualitative information, e.g. an ordinal relation allowing for an intuitive interpretation of the value (too low, too high or okay) \cite{de1984qualitative}. Hence, we use information usually coming from quality and safety requirements, physical limitations or system specifications defining valid and invalid intervals on states \cite{iec61360}.
	This information is integrated into FOL using 
	\begin{equation} \label{eq:assign}
		\begin{aligned}
			x_i < lb_i &\rightarrow low(x_i), \\
			lb_i \leq x_i \leq ub_i &\rightarrow ok(x_i), \\
			x_i > ub_i &\rightarrow high(x_i),
		\end{aligned}
	\end{equation}
	where $lb_i, ub_i \in \mathbb{R}$ define the interval bounds.
	
	Existing approaches on reconfiguration directly encode \eqref{eq:assign} using Satisfiability Modulo Theories \cite{Balzereit2020}. However, this increases the complexity of problem solving since in addition to a SAT problem, Linear Arithmetic needs to be handled. Here, the truth values of the constraints is evaluated first, which is possible since $x_i$, $lb_i$ and $ub_i$ are known. Thus, an observation $\alpha_X$ implies a truth assignment to the predicates in \eqref{eq:assign}, denoted by $\alpha_X^\mathrm{int}$.

	\subsection{Reconfiguration System Model} \label{sec:systemModel}
	
	The reconfiguration system model $SM$ consists of a set of constraints in FOL which together satisfy \eqref{eq:valid}, \eqref{eq:invalid}, and \eqref{eq:operations}. 
	Therefore, for each invalid search node, the operations recovering a valid search node are identified and a constraint $s_i \rightarrow \varphi_i(O)$, where $\varphi_i(O)$ describes the operations, is created.
	Since operations change the value of the binary inputs $B$, $\varphi_i(O)$ can be replaced by a formula $\tilde{\varphi_i}(B)$. The implication $s_i \rightarrow \tilde{\varphi_i}(B)$ contains a contradiction if the search node $s_i$, and thus the configuration $\langle \alpha_{X_i}, \alpha_{B_i} \rangle$ is invalid.

	Using the discretization from step \textbf{I}, these constraints can also be written for more than one search node: E.g. the constraint $low(x_i) \rightarrow \tilde{\varphi^*_i}(B)$
	defines the reconfiguration operations for all nodes $s_i$ for that $low(x_i)$ holds. However, the approach is compatible with any discretization $g(x_1, x_2, ...)$ assigning arbitrary predicates $p_1, p_2, ...$ truth values, as long as the formula $\tilde{\varphi^*_i}(B)$ on the inputs recovering validity can be identified, so $p_i \rightarrow \tilde{\varphi^*_i}(B)$.
	The conjunction of these implications for every predicate $p_i$ representing discretized areas forms $SM$.

	The Three-Tank system presented by Blanke et al. \shortcite{blanke2006diagnosis} serves as example. The system consists of three tanks $T_1, T_2, T_3$. Every tank has an inflow controlled by the pumps $p_1, p_2$. The valves $v_{12}^a, v_{23}^a$ and $v_{12}^b, v_{23}^b$ are located at a height of 30 and 0 cm, respectively. Thus, the flows depend on the water levels in the tanks. The goal of the system is to supply a constant outflow of $T_2$ by maintaining the water level between 10 and 20 cm.
	
	Here, we assume that $\bigwedge_{x_i \in X} ok(x_i)$ ensures reachability of the system goal, so that in case of no faults all states are in their valid intervals and the system goal is reachable \cite{Vaennman2007}. So, every configuration leading to all states being ok is valid. 
	In case of a leak in $T_1$, the level becomes too low. The system can be reconfigured by exchanging $T_1$, described by $b_{T_1}^\mathrm{EXT} \in B^\mathrm{EXT}$, or closing valve $v_{12}^b$. In FOL, this is expressed by $low(x_1) \rightarrow \neg b_{12}^b \vee b_{T_1}^\mathrm{EXT}$. A configuration not satisfying this constraint is invalid.
	
	\subsection{Overall Solution} \label{sec:overallSolution}
	
	\begin{algorithm}
		\small
		\caption{SATReconf}
		\label{alg:reconf}
		\KwIn{$HRP = (SM, s_i = \langle \alpha_X^\mathrm{int}, \alpha_B  \rangle)$}
		\KwOut{$\alpha_B^*$ or \textit{error}}
		$n^\mathrm{ub} := 1$ \\ \label{line:initUB}
		\While{$n^\mathrm{ub} \leq \mathrm{|B|}$}{ \label{line:while} 
			$\beta := (\sum_{k} \big( \alpha_{B_k} \oplus \alpha_{B_k}^* \big)  \leq n^\mathrm{ub})$ \label{line:addUB} \\
			\eIf{$\mathrm{SAT}(SM \wedge \alpha_X^\mathrm{int} \wedge \beta)$ \label{line:ifNewCheckConf}}{
				$\alpha_B^* := $ ASSIGN($SM \wedge \alpha_X^\mathrm{int} \wedge \beta$) \label{line:modelCheckConf}\\
				\Return{$\alpha_B^*$} \label{line:returnVCheckConf}
			}{ \label{line:else2CheckConf}
				$n^\mathrm{ub} := n^\mathrm{ub} + 1$ \label{line:incUB} \\
			}
		}
		\Return{error} \label{line:error}
	\end{algorithm}
	
	Algorithm 1 shows how a new configuration is determined. The reconfiguration system model $SM$ and the search node $s_i$ are input to the algorithm. The SAT solver contains functions SAT and ASSIGN, which check if the given formula is satisfiable and return the corresponding assignment, respectively. The core of the algorithm is the satisfiability check of the formula in line \ref{line:ifNewCheckConf} which consists of the system model $SM$ and the discretized observation of states $\alpha_X^\mathrm{int}$.
	To ensure a minimum of reconfiguration operations is chosen, the constraint $\beta$ contains an upper bound for binary changes $n^\mathrm{ub}$ (l. \ref{line:addUB}). Therefore, the observation $\alpha_B$ is compared to the searched assignment $\alpha_{B}^*$: Every single assignment $\alpha_{B_k}$ is compared to $\alpha_{B_k}^*$, and a cardinality constraints ensures that a reconfiguration of minimal cardinality, i.e. the reconfiguration requiring the least operations, is identified. Starting with 1 (l. \ref{line:initUB}), the upper bound is increased (l. \ref{line:incUB}) until a satisfying assignment to the formula is found or the number of possible changes, given by the number of binary variables, is reached (l. \ref{line:while}).
	If a satisfying assignment exists (l. \ref{line:ifNewCheckConf}), it is identified (l. \ref{line:modelCheckConf}) and returned (l. \ref{line:returnVCheckConf}). The output $\alpha_B^*$ satisfies $SM \wedge s_i^* \not\models \bot$ with $s_i^* = \langle  \alpha_X^\mathrm{int}, \alpha_B^* \rangle$. If no assignment can be found, no reconfiguration exists and an error is returned (l. \ref{line:error}).
	
	\section{Experimental Results}
	
	For the experimental results, we use the Three-Tank System and a Two-Tank system which has already been used for reconfiguration purposes \cite{balzereit2021Gradient}. The system consists of two tanks $T_1, T_2$. Every tank has an inflow and an outflow controlled by the valves $v_{01}, v_{10}, v_{02}, v_{20}$. The pumps $p_{12}, p_{21}$ connect the tanks. Level and temperature are measured in each tank. The system goal is to supply water between 65 and 75$^{\circ}$C through $v_{10}$ and between 10 and 20$^{\circ}$C through $v_{20}$ while keeping the level in each tank between 30 and 40 cm.
	
	Tank systems are hybrid systems since they contain discrete variables such as opening states of valves and continuous variables such as water levels (see \textit{RQ2}) \cite{khorasgani2019mode}.
	The systems are simulated in Modelica \cite{Elmqvist1998}, the satisfiability of the logical formula is checked using the solver Z3 \cite{DeMoura2008}.
	For each system, one spare tank, which can replace a faulty tank, exists.
	
	For reconfiguration, faults are not separated according to the affected component but to their effect on the states. Continuous faults lead to a continuous change of at least one state, for example a leak in a tank leads to a continuous decrease of the water level. Discrete faults lead to an abrupt change of at least one state, for example an abrupt addition of water to a tank leads to an abrupt increase of the water level. Temporal effects like a creeping loss of temperature are handled as soon as a state reaches an invalid interval bound. Faults which have no effects on the states are not within the scope of this article.

	\paragraph{Empirical Evaluation}

	The Two-Tank system encountered 58 faults, the Three-Tank system 39.
	In both systems, the scenarios cover faults of every component and connection. In addition to single faults, multiple fault scenarios consisting of two continuous, one continuous and one discrete, or two discrete faults are simulated.
	The results of the evaluation are shown in Table \ref{tab:results}. All of the continuous, discrete and multiple continuous faults are reconfigured correctly. When it comes to multiple faults with at least one discrete fault, the method is only capable of handling 65\% (multiple continuous + discrete) to 57\% (multiple discrete) of faults since multiple faults often lead to conflicts: Multiple states cause the invalid configuration which may lead to contradicting constraints.
	
	\begin{table}
		\small
		\begin{tabular}{lccc}
			\hline
			\textbf{Kind of Faults}             & \textbf{\# cases} & \textbf{reconf.} & \textbf{in \%} \\ \hline \hline
			\textbf{Two-Tank System}            & \textbf{58}       & \textbf{50}                                                              & \textbf{86}                \\ \hline
			\textbf{continuous}                 & \textbf{16}       & \textbf{16}                                                              & \textbf{100}   \\
			leak in one tank                    & 2                 & 2                                                                        &                \\
			valve stuck open                    & 4                 & 4                                                                        &                \\
			valve stuck close                   & 4                 & 4                                                                        &                \\
			pump stuck full                     & 2                 & 2                                                                        &                \\
			pump blocked                        & 2                 & 2                                                                        &                \\
			continuous temperature loss/rise    & 2                 & 2                                                                        &                \\ \hline
			\textbf{discrete}                   & \textbf{22}       & \textbf{22}                                                              & \textbf{100}   \\
			drop in wl (20\%-70\%)              & 5                 & 5                                                                        &                \\
			rise in wl (20\%-70\%)              & 5                 & 5                                                                        &                \\
			drop in temperature (7\%-42\%)      & 6                 & 6                                                                        &                \\
			rise in temperature (7\%-42\%)      & 6                 & 6                                                                        &                \\ \hline
			\textbf{multiple continuous}        & \textbf{4}        & \textbf{4}                                                               & \textbf{100}   \\
			heating failure + valve stuck       & 2                 & 2                                                                        &                \\
			cooler failure + valve stuck        & 2                 & 2                                                                        &                \\ \hline
			\textbf{multiple continuous + discrete}            & \textbf{10}       & \textbf{5}                                                               & \textbf{50}    \\
			heating failure + drop in wl        & 5                 & 5                                                                        &                \\
			cooler failure + rise in wl         & 5                 & 0                                                                        &                \\ \hline
			\textbf{multiple discrete}          & \textbf{6}        & \textbf{3}                                                               & \textbf{50}    \\
			drop in wl +  in temperature        & 5                 & 2                                                                        &                \\
			drop in wl in 2 tanks               & 1                 & 1                                                                        &                \\ \hline \hline
			\textbf{Three-Tank System}          & \textbf{39}       & \textbf{37}                                                              & \textbf{95}    \\ \hline
			\textbf{continuous}                 & \textbf{14}       & \textbf{14}                                                              & \textbf{100}   \\
			leak in one tank                    & 2                 & 2                                                                        &                \\
			valve stuck open                    & 4                 & 4                                                                        &                \\
			valve stuck close                   & 4                 & 4                                                                        &                \\
			pump stuck full power               & 2                 & 2                                                                        &                \\
			pump blocked                        & 2                 & 2                                                                        &                \\ \hline
			\textbf{discrete}                   & \textbf{10}       & \textbf{10}                                                              & \textbf{100}   \\
			drop in wl (20\%-70\%)           & 5                 & 5                                                                        &                \\
			rise in wl  (20\%-70\%)           & 5                 & 5                                                                        &                \\ \hline
			\textbf{multiple continuous}           & \textbf{4}        & \textbf{4}                                                               & \textbf{100}   \\
			leak in one tank + valve stuck      & 4                 & 4                                                                        &                \\ \hline
			\textbf{multiple continuous + discrete}               & \textbf{10}       & \textbf{8}                                                               & \textbf{80}    \\
			valve stuck open + drop in wl     & 5                 & 4                                                                        &                \\
			pump blocked + rise in wl        & 5                 & 4                                                                        &                \\ \hline
			\textbf{multiple discrete}             & \textbf{1}        & \textbf{1}                                                               & \textbf{100}   \\
			drop in wl  in 2 tanks            & 1                 & 1                                                                        &                \\ \hline
		\end{tabular}
		\caption{For each kind of faults listed in col. 1 (wl = water level), the number of simulated cases is listed in col. 2. The number of correctly reconfigured faults is listed in col. 3, in percent in col. 4.}
		\label{tab:results}
	\end{table}
	
	\paragraph{Runtime of SATReconf}
	
	SATReconf searches a solution to an NP-complete problem such that the runtime for the worst case rises exponentially with the number of binary inputs. However, a SAT solver often can find a solution in a shorter time \cite{perlovsky1998conundrum}.
	
	\section{Conclusion}
	
	Hybrid systems are prone to faults, which often lead to the system goal being no longer reachable. Reconfiguration is the task of setting input variables and exchanging physical components to recover the system's ability to reach its goal. Especially in the case of hybrid systems, the problem of reconfiguration is not yet solved. In this article, a novel approach to reconfiguration of hybrid systems is presented. Therefore, reconfiguration is formalized as a search problem which enables independence of the modeling formalism of the hybrid system (see \textit{RQ1}). The originally large search space consisting of valid and invalid configurations is reduced using a discretization of continuous variables. The reachability of the system goal is encoded in terms of valid and invalid search nodes. The search is encoded in SAT which allows handling the binary concept of validity and efficient solving by leveraging on many years of SAT research (see \textit{RQ2}). Therefore, a logical formula representing the search is created; a SAT solver is used to find such an assignment of binary input variables that change the invalid search node to a valid one. This solution approach relies on observations of the system variables only, requiring no information about faults.
	
	Experimental results show that the solution approach is capable of handling single faults and even multiple continuous faults. Multiple discrete faults are difficult to handle since they often lead to conflicts. Adding a prioritization to the states will enable the algorithm to handle such faults one by one. Since the simulation scenarios are designed to introduce faults of every component and connection, as well as combinations of those, we believe that a wide range of possible faults is covered. And since no assumption about process engineering is made, we expect the approach to be transferable to hybrid systems from other application areas. To corroborate these assumptions, we will evaluate the approach using examples from other areas, including real-world systems. 
	Since the search at the moment is encoded as boolean satisfiability an arbitrary valid solution is identified. In the future, we will extend the approach to find the optimal solution.
	
	\bibliographystyle{unsrt}
	\bibliography{locBib}

\end{document}